\documentclass[10pt,twocolumn,letterpaper]{article}

\usepackage{cvpr}              %

\usepackage{graphicx}
\usepackage{amsmath}
\usepackage{amssymb}
\usepackage{booktabs}

\usepackage[pagebackref,breaklinks,colorlinks]{hyperref}

\usepackage[capitalize]{cleveref}
\crefname{section}{Sec.}{Secs.}
\Crefname{section}{Section}{Sections}
\Crefname{table}{Table}{Tables}
\crefname{table}{Tab.}{Tabs.}

\newcommand{\myparagraph}[1]{\vspace{0.1em}\noindent\textbf{#1}}

\def\cvprPaperID{8665} %
\def\confName{CVPR}
\def\confYear{2023}

\begin{document}

\title{Instant-NVR: Instant Neural Volumetric Rendering for Human-object Interactions from Monocular RGBD Stream}

\author{Yuheng Jiang$^{1,2}$* \;\, Kaixin Yao$^{1,2}$* \;\, Zhuo Su$^{3}$ \;\, Zhehao Shen$^{1}$ \;\, Haimin Luo$^{1}$ \;\, Lan Xu$^{1}$} 

\makeatletter
\let\@oldmaketitle\@maketitle%
\renewcommand{\@maketitle}{
	\@oldmaketitle%
	\centering
	\vspace{-8mm}
	{\large \textsuperscript{1}ShanghaiTech University}\quad \quad
	{\large \textsuperscript{2}NeuDim}\quad \quad
	{\large \textsuperscript{3}Pico IDL, ByteDance}\quad \quad

	\vspace{8mm}
}
\makeatother

\maketitle

\def\thefootnote{*}\footnotetext{Equal Contribution.}

\begin{abstract}
  Convenient 4D modeling of human-object interactions is essential for numerous applications. However, monocular tracking and rendering of complex interaction scenarios remain challenging.
In this paper, we propose Instant-NVR, a neural approach for instant volumetric human-object tracking and rendering using a single RGBD camera.
It bridges traditional non-rigid tracking with recent instant radiance field techniques via a multi-thread tracking-rendering mechanism.
In the tracking front-end, we adopt a robust human-object capture scheme to provide sufficient motion priors.
We further introduce a separated instant neural representation with a novel hybrid deformation module for the interacting scene.
We also provide an on-the-fly reconstruction scheme of the dynamic/static radiance fields via efficient motion-prior searching.
Moreover, we introduce an online key frame selection scheme and a rendering-aware refinement strategy to significantly improve the appearance details for online novel-view synthesis.
Extensive experiments demonstrate the effectiveness and efficiency of our approach for the instant generation of human-object radiance fields on the fly, notably achieving real-time photo-realistic novel view synthesis under complex human-object interactions. Project page: \href{https://nowheretrix.github.io/Instant-NVR/}{https://nowheretrix.github.io/Instant-NVR/}.

\end{abstract}

    \section{Introduction}
The accurate tracking and photo-realistic rendering for human-object interactions are critical for numerous human-centric applications like telepresence, tele-education or immersive experience in VR/AR. 
However, a convenient solution from monocular input, especially for on-the-fly setting, remains extremely challenging in the vision community.

\begin{figure}[tbp] 
	\centering 
	\includegraphics[width=1\linewidth]{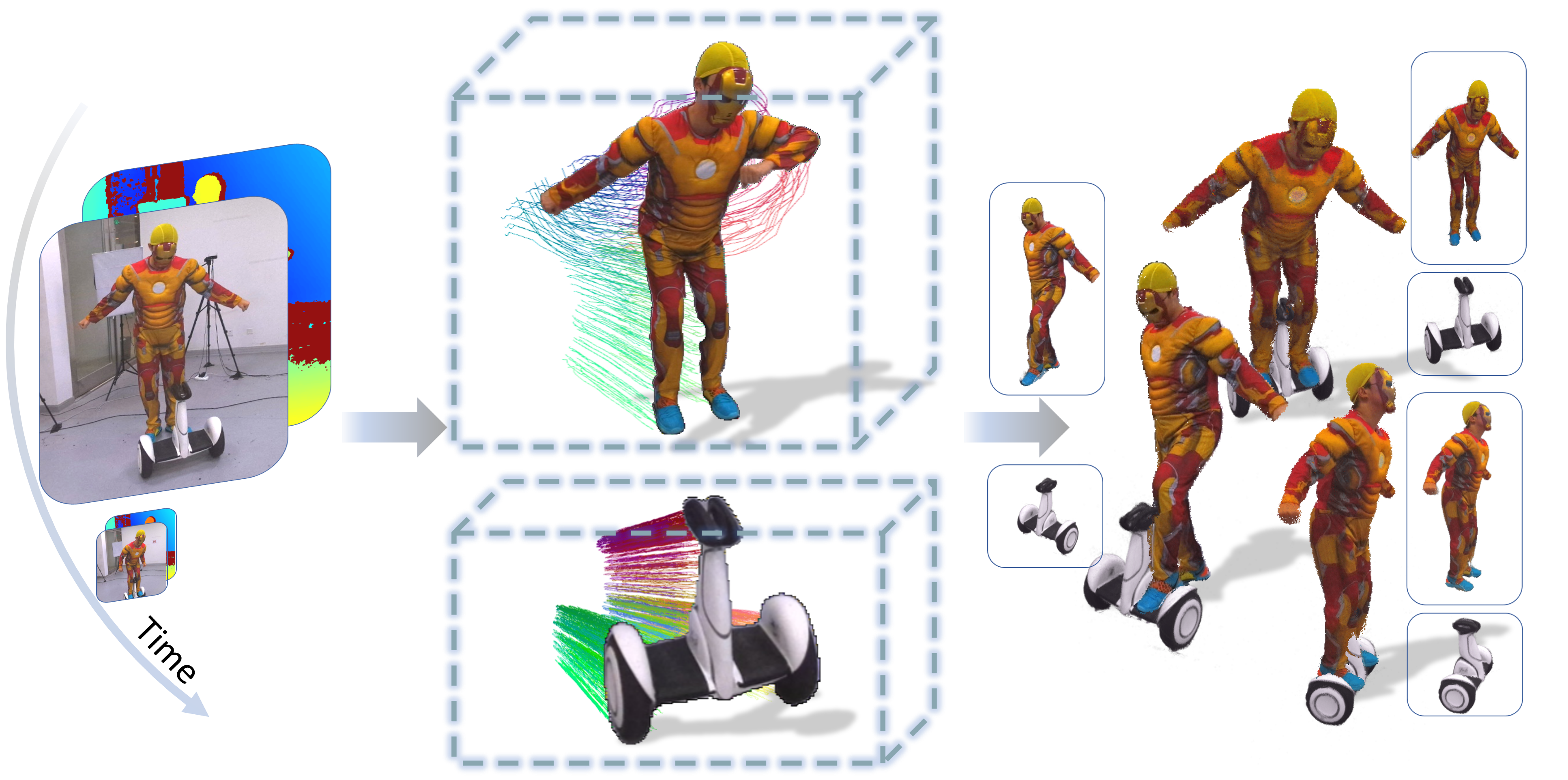} 
	\vspace{-20pt} 
	\caption{Our Instant-NVR adopts a separated instant neural representation to achieve photo-realistic rendering for human-object interacting scenarios.} 
	\label{fig:fig_1_teaser} 
	\vspace{-15pt} 
\end{figure}

Early high-end solutions~\cite{bradley2008markerless,collet2015high,TotalCapture,guo2019relightables} require dense cameras for high-fidelity reconstruction. Recent approaches~\cite{dou2016fusion4d,UnstructureLan,motion2fusion,yu2021function4d,jiang2022neuralhofusion, suo2021neuralhumanfvv, sun2021HOI-FVV} need less RGB or RGBD video inputs (from 3 to 8 views) by using volumetric tracking techniques~\cite{li2009robust,newcombe2015dynamicfusion}. Yet, the multi-view setting is still undesirable for consumer-level daily usage.
Differently, the monocular method with a single handiest commercial RGBD camera is more practical and attractive.
For monocular human-object modeling, most approaches~\cite{zhang2020object,zhang2020phosa,bhatnagar22behave,xie2022chore,haresh2022articulated,wang2022reconstruction} track the rigid and skeletal motions of object and human using a pre-scanned template or parametric model. Besides, the monocular volumetric methods~\cite{newcombe2015dynamicfusion,KillingFusion2017cvpr,FlyFusion,DoubleFusion,robustfusion} obtain detailed geometry through depth fusion, while the recent advance~\cite{su2022robustfusionPlus} further extends it into the human-object setting. However,  they fail to generate realistic appearance results, restricted by the limited geometry resolution.

Recent neural rendering advances, represented by Neural Radiance Fields (NeRF)~\cite{nerf}, have recently enabled photo-realistic rendering with dense-view supervision. Notably, some recent dynamic variants of NeRF~\cite{li2021neural,weng_humannerf_2022_cvpr,tretschk2021nonrigid,yoon2022learning,luo2022artemis,wang2022fourier,zhao2022human} obtain the compelling novel-view synthesis of human activities even under monocular capturing. However, they rely on tedious and time-consuming per-scene training to fuse the temporal observations into the canonical space, thus unsuitable for on-the-fly usage like telepresence. 
Only recently, Instant-NGP~\cite{muller2022instant} enables fast radiance field generation in seconds, bringing the possibility for on-the-fly radiance field modeling. Yet, the original Instant-NGP can only handle static scenes. Few researchers explore the on-the-fly neural rendering strategies for human-object interactions, especially for monocular setting.

In this paper, we present \textit{Instant-NVR} -- an instant neural volumetric rendering system for human-object interacting scenes using a single RGBD camera. As shown in  Fig.~\ref{fig:fig_1_teaser}, Instant-NVR enables instant photo-realistic novel view synthesis via on-the-fly generation of the radiance fields for both the rigid object and dynamic human.
Our key idea is to bridge the traditional volumetric non-rigid tracking with instant radiance field techniques. Analogous to the tracking-mapping design in SLAM, we adopt a multi-thread and tracking-rendering mechanism. 
The tracking front-end provides online motion estimations of both the performer and object, while the rendering back-end reconstructs the radiance fields of the interaction scene to provide instant novel view synthesis with photo-realism. 

For the tracking front-end, we first utilize off-the-shelf instant segmentation to distinguish the human and object from the input RGBD stream. Then, we adopt an efficient non-rigid tracking scheme for both the performer and rigid object, where we adopt both embedded deformation~\cite{sumner2007embedded} and SMPL~\cite{SMPL2015} to model human motions. 
For the rendering back-end, inspired by Instant-NGP~\cite{muller2022instant} we adopt a separate instant neural representation. Specifically, both the dynamic performer and static object are represented as implicit radiance fields with multi-scale feature hashing in the canonical space and share volumetric rendering for novel view synthesis. For the dynamic human, we further introduce a hybrid deformation module to efficiently utilize the non-rigid motion priors.
Then, we modify the training process of radiance fields into a key-frame based setting, so as to enable graduate and on-the-fly optimization of the radiance fields within the rendering thread. For the dynamic one, we further propose to accelerate our hybrid deform module with a hierarchical and GPU-friendly strategy for motion-prior searching.
Yet we observe that naively selecting key-frames with fixed time intervals will cause non-evenly distribution of the captured regions of the dynamic scene. It results in unbalanced radiance field optimization and severe appearance artifacts during free-view rendering. To that end, we propose an online key-frame selection scheme with a rendering-aware refinement strategy. It jointly considers the visibility and motion distribution across the selected key-frames, achieving real-time and photo-realistic novel-view synthesis for human-object interactions.

To summarize, our main contributions include:
\begin{itemize} 
	\setlength\itemsep{0em}
	
	\item We present the first instant neural rendering system under human-object interactions from an RGBD sensor.

	\item We introduce an on-the-fly reconstruction scheme for dynamic/static radiance fields using the motion priors through a tracking-rendering mechanism.
	
	\item We introduce an online key frame selection scheme and a rendering-aware refinement strategy to significantly improve the online novel-view synthesis.

\end{itemize}

    \begin{figure*}[t] 
	\begin{center} 
		\includegraphics[width=\linewidth]{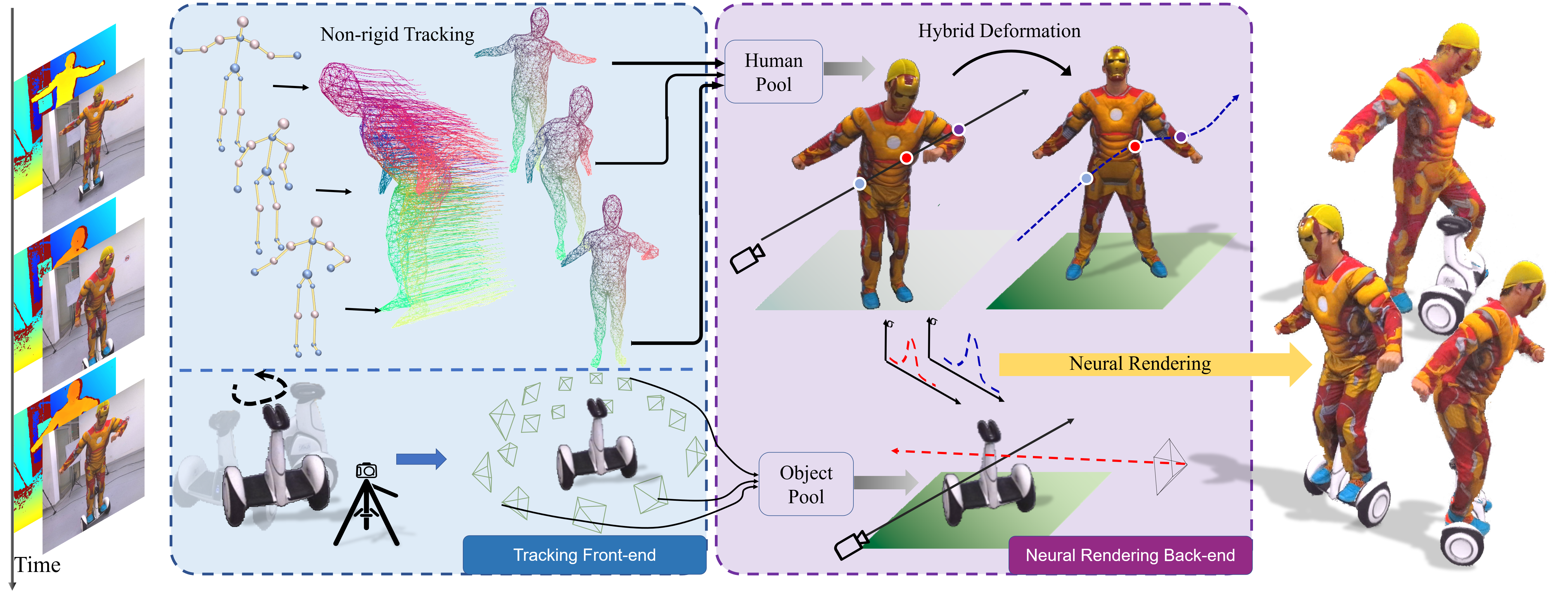} 
	\end{center} 
    \vspace{-20pt}
	\caption{Our approach consists of two stages. The tracking front-end (Sec.~\ref{sec:Tracker}) captures human and object motions, while the rendering back-end (Sec.~\ref{sec:Renderer}) separately reconstructs the human-object radiance fields on-the-fly, for instant novel view synthesis with photo-realism.} 
	\label{fig:fig_2_overview} 
	\vspace{-5pt}
\end{figure*}

\section{Related Work} 
\noindent{\textbf{Traditional Human Volumetric Capture.}} 
Human volumetric capture and reconstruction have been widely investigated to achieve detailed geometry reconstruction and accurate tracking.
A series of works are proposed to make volumetric fusion more robust with SIFT features~\cite{innmann2016volumedeform}, multi-view systems~\cite{dou2016fusion4d,motion2fusion},
scene flow~\cite{Wang_2020_WACV},
human articulated skeleton prior~\cite{DoubleFusion,BodyFusion}, extra IMU sensors~\cite{HybridFusion}, data-driven prior~\cite{robustfusion, su2022robustfusionPlus }, learned correspondences~\cite{bozic2020deepdeform}, neural deformation graph~\cite{bozic2021neural, lin2022occlusionfusion} or implicit function~\cite{yu2021function4d,jiang2022neuralhofusion}.
Starting from the pioneering work DynamicFusion~\cite{newcombe2015dynamicfusion} which benefits from the GPU solvers, the high-end solutions~\cite{dou2016fusion4d,motion2fusion} rely on the multi-view camera system and complex calibration.
VolumeDeform~\cite{innmann2016volumedeform} combines depth-based correspondences with sparse SIFT features to reduce drift.
KillingFusion~\cite{KillingFusion2017cvpr} and SobolevFusion~\cite{slavcheva2018sobolevfusion} support topology changes via more constraints on the motion fields. 
Thanks to the human parametric model~\cite{SMPL2015}, DoubleFusion~\cite{DoubleFusion} proposes the two-layer representation to capture scene more robustly. UnstructuredFusion~\cite{UnstructureLan} extends it to an unstructured multi-view setup. RobustFusion~\cite{su2022robustfusionPlus} further handles the challenging human-object interaction scenarios. Besides, Function4d~\cite{yu2021function4d} and NeuralHOFusion~\cite{jiang2022neuralhofusion} marry the non-rigid-tracking with implicit modeling. However, these methods are dedicated to getting detailed geometry without focusing on high-quality texture and most methods can not handle human-object interactions. Comparably, our approach bridges the traditional volumetric capture and neural rendering advances, achieving photo-realistic rendering results under human-object interactions.

\noindent{\textbf{Static Neural Scene Representations.}}
Coordinates-based neural scene representations in static scenes produce impressive novel view synthesis results and show huge potential.
Various data representations are adopted to obtain better performance and characteristics, such as point-clouds~\cite{Wu_2020_CVPR,aliev2019neural,suo2020neural3d}, voxels~\cite{lombardi2019neural}, textured meshes~\cite{thies2019deferred,liu2019neural}, occupancy~\cite{oechsle2021unisurf, shao2022doublefield} or SDF~\cite{wang2021neus, DeepSDF}.
Meanwhile, Since the vanilla NeRF which requires hours of training is time-consuming, some NeRF extensions~\cite{yu2021plenoctrees,yu_and_fridovichkeil2021plenoxels,muller2022instant } are proposed to accelerate both training and rendering. Plenoctrees~\cite{yu2021plenoctrees} utilizes the octree to skip the empty regions. Plenoxels\cite{yu_and_fridovichkeil2021plenoxels} parameterizes the encoding using spherical harmonics on the explicit 3D volume. Instant-NGP~\cite{muller2022instant} utilizes the multi-scale feature hashing and TCNN to speed up. Though its rendering speed seems possible to train on-the-fly, they do not have a specific design for streaming input and only can recover static scenes. Comparably, our Instant-NVR achieves on-the-fly efficiency based on the Instant-NGP~\cite{muller2022instant}.

\noindent{\textbf{Dynamic Neural Scene Representations.}} Novel view synthesis in  dynamic scenes is an important research problem.
D-NeRF~\cite{pumarola2020d} and Non-rigid NeRF~\cite{tretschk2021nonrigid} leverage the displacement field to represent the motion while Nerfies~\cite{park2021nerfies} and HyperNeRF~\cite{park2021hypernerf} use the SE(3) field. Moreover, some researchers focus on human reconstruction and utilize the human prior. Neuralbody anchors latent code on the SMPL~\cite{SMPL2015} vertices. Humannerf ~\cite{zhao2022humannerf} combines the SMPL warping and deformation net to construct the motion field. TAVA~\cite{li2022tava} learns the skinning weight for joints via root-finding and can generalize to novel pose. DeVRF~\cite{liu2022devrf} incorporates 4D-motion volume into the NeRF pipeline. NDR~\cite{Cai2022NDR} defines a bijective function which naturally compatible with the cycle consistency. However, most methods rely on multi-view camera input and the training is costly. Comparably, our Instant-NVR bridges the non-rigid volumetric capture with the instant radiance field training, achieving photo-realistic rendering results from monocular RGBD stream.

    \begin{figure*}[t] 
	\begin{center} 
		\includegraphics[width=1\linewidth]{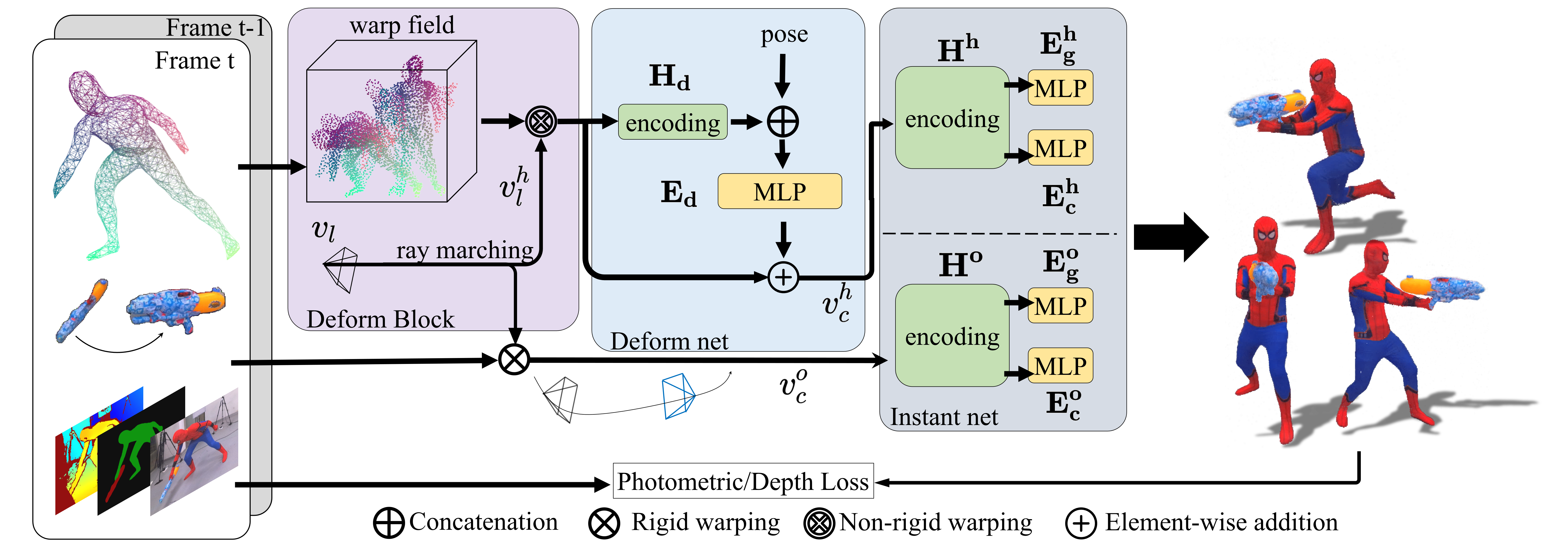}
	\end{center} 
	\vspace{-5mm}
	\caption{Our Neural Rendering back-end adopts a separated neural representation. Left are the input RGBD images with motions. Middle is a separate rendering engine that includes a hybrid deformation module and volumetric rendering. Right are rendering results.} 
	\label{fig:renderer} 
	\vspace{-5pt}
\end{figure*} 
\section{Overview}
From monocular RGBD input, Instant-NVR bridges the real-time non-rigid capture with instant neural rendering, allowing for high-quality novel-view synthesis under human-object interactions. 
As illustrated in Fig.~\ref{fig:fig_2_overview}, our system consists of two cooperating threads: a tracking front-end (Sec.~\ref{sec:Tracker}) and a neural rendering back-end (Sec.~\ref{sec:Renderer}).

\vspace{2mm}\noindent{\bf Tracking Front-end.}
We extend the traditional volumetric tracking ~\cite{newcombe2015dynamicfusion,DoubleFusion, UnstructureLan} into a human-object setting. For non-rigid human capture, we adopt the embedded deformation(ED)~\cite{sumner2007embedded} and SMPL~\cite{SMPL2015} as motion representations. For object, we directly track its rigid motions via the Iterative Closest Point(ICP) algorithm. This thread provides accurate per-frame human-object motion priors, enabling the integration of all the radiance information into the global canonical space.
Note that the reconstructed volumetric geometry suffers from discrete and low-resolution artifacts. 
Thus, we only transmit the motion priors with the RGBD images to the rendering thread and discard the explicit volumetric geometry prior.

\vspace{2mm}\noindent{\bf Neural Rendering Back-end.} 
We extend instant radiance fields~\cite{muller2022instant} to the monocular and dynamic human-centric scenes, where we maintain the canonical instant radiance fields for both dynamic human and rigid objects separately. We introduce a lightweight pose-conditioned deformation module to learn the residual motion to refine the initial warping provided by motion priors. 
To enable on-the-fly radiance field generation and rendering, we adapt the training process into a key frame setting with the aid of efficient motion-prior caching.
We introduce a key frame selection method to jointly consider the diversity of capturing view and human pose, visibility maps, and the input image quality. We further adaptively refine the appearance output in the rendering view with more analogous spatial-temporal capturing views.
Note that our rendering thread reconstructs the radiance fields online to provide instant novel view synthesis with photo realism.

\section{Method}\label{sec:algorithm}

\subsection{Tracking Front-end} \label{sec:Tracker}

\noindent{\bf Human Non-rigid Tracking.}
We follow the traditional volumetric capture methods~\cite{DoubleFusion, su2022robustfusionPlus} to track human non-rigid motions. 
Specifically, we parameterize human non-rigid motions as an embedded deformation graph $W=\{dq_i,x_i\}$, where $x_i$ is the coordinates of the sampled ED node in canonical space and $dq_i$ is the dual quaternions representing the corresponding rigid transformation in $SE(3)$ space.
Each 3D point $v_c$ in the canonical space can be wrapped into the live space using an efficient and accurate motion interpolation method Dual-Quaternion Blending ($DQB$):
\begin{equation}
\begin{split}
\begin{aligned}
    {DQB}(v_c) = \sum_{i \in \mathcal{N}(v_c)} w(x_i,v_c)dq_i, \\
    \tilde v_c = SE3( {DQB}  (v_c)) v_c.
\end{aligned}
\end{split}
\label{equ2}
\end{equation}
where $\mathcal{N}(v_c)$ is a set of neighboring ED nodes of $v_c$,  $w(x_i,v_c)$ is the influence weight of the $i-th$ node $x_i$ to $v_c$ and formulated as $w(x_i,v_c) = exp( -\left\|\mathbf{v}_c-\mathbf{x}_i\right\|_2^2 /r^2)$. $r$ is the influence radius (0.1 in our setting).
Note that the ED-only-based human tracking is fragile since 
the non-rigid ICP often fails at fast articulated human motions due to losing correspondence.
Therefore, we also introduce the SMPL inner body with shape parameters ${\beta}$ and pose parameters ${\theta}$ as the skeleton prior and utilize ${\theta}$ with the skinning weight to wrap 3D point $v_c$, which further constrain the ED motion tracking within a reasonable motion scale.
Please refer to \cite{DoubleFusion} for details about the ED-sampling and double layer motion representation.

To calculate the final ED-based motion, we jointly optimize the skeleton pose ${\theta}$ and ED non-rigid motion field $W$ as follows:
\begin{equation}\label{eq:Emot_track}
\begin{split}
E(W,{\theta})=&\lambda_{\mathrm{data}}E_{\mathrm{data}}+\lambda_{\mathrm{bind}}E_{\mathrm{bind}}+\lambda_{\mathrm{reg}}E_{\mathrm{reg}}+\\
&\lambda_{\mathrm{prior}}E_{\mathrm{prior}} + \lambda_{\mathrm{pose}}E_{\mathrm{pose}}+ \lambda_{\mathrm{inter}}E_{\mathrm{inter}}.
\end{split}
\end{equation}
The data term $E_{\mathrm{data}}$ measures the point-to-plane distances between the deformed model and the current input depth map:
\begin{equation}\label{eq:Eemot_data}
\vspace{-2pt}
\boldsymbol{E}_{\mathrm{data}}=\sum\limits_{(\mathbf{v}_c,\mathbf{u})\in{\mathcal{P}}}{\psi(\textbf{n}_{{\mathbf{u}}}^{T}(\tilde{\mathbf{v}}_c-\mathbf{u}))},
\end{equation}
where $\mathbf{u}$ is a sampled point in the depth map, $\mathbf{n}_{\mathbf{u}}$ is its normal, and $\mathbf{v}_c$ denotes its closest point on the fused surface. $\mathcal{P}_i$ is the set of correspondences found via a projective local search \cite{newcombe2015dynamicfusion}.
Besides, the binding term $E_{\mathrm{bind}}$ constrains both skeleton and final ED motions to be consistent while the geometry regularity term $E_{\mathrm{reg}}$ produces locally as-rigid-as-possible (ARAP) motions to prevent overfitting to depth inputs. These two terms are detailed in \cite{DoubleFusion, guo2017real}.
The pose prior term $E_{\mathrm{prior}}$ from \cite{keepitSMPL} penalizes the unnatural poses. 
Both the pose term $E_{\mathrm{pose}}$ and interaction term $E_{\mathrm{inter}}$ are form~\cite{su2022robustfusionPlus} to encourage natural motion capture during human-object interactions.
Note that the optimization non-linear least squares problem in Eqn. \ref{eq:Emot_track} is solved using LM method with the PCG solver on GPU~\cite{guo2017real,dou2016fusion4d}.

\noindent{\bf Object Rigid Tracking.}
For rigid tracking of objects, we follow \cite{su2022robustfusionPlus} to optimize the rigid motions and transform them to camera pose $T^t$ under the ICP framework, in which we fuse the depth map to a canonical TSDF volume to maintain the stable correspondence for robust object tracking.

\begin{figure}[t] 
	\begin{center} 
		\includegraphics[width=1\linewidth]{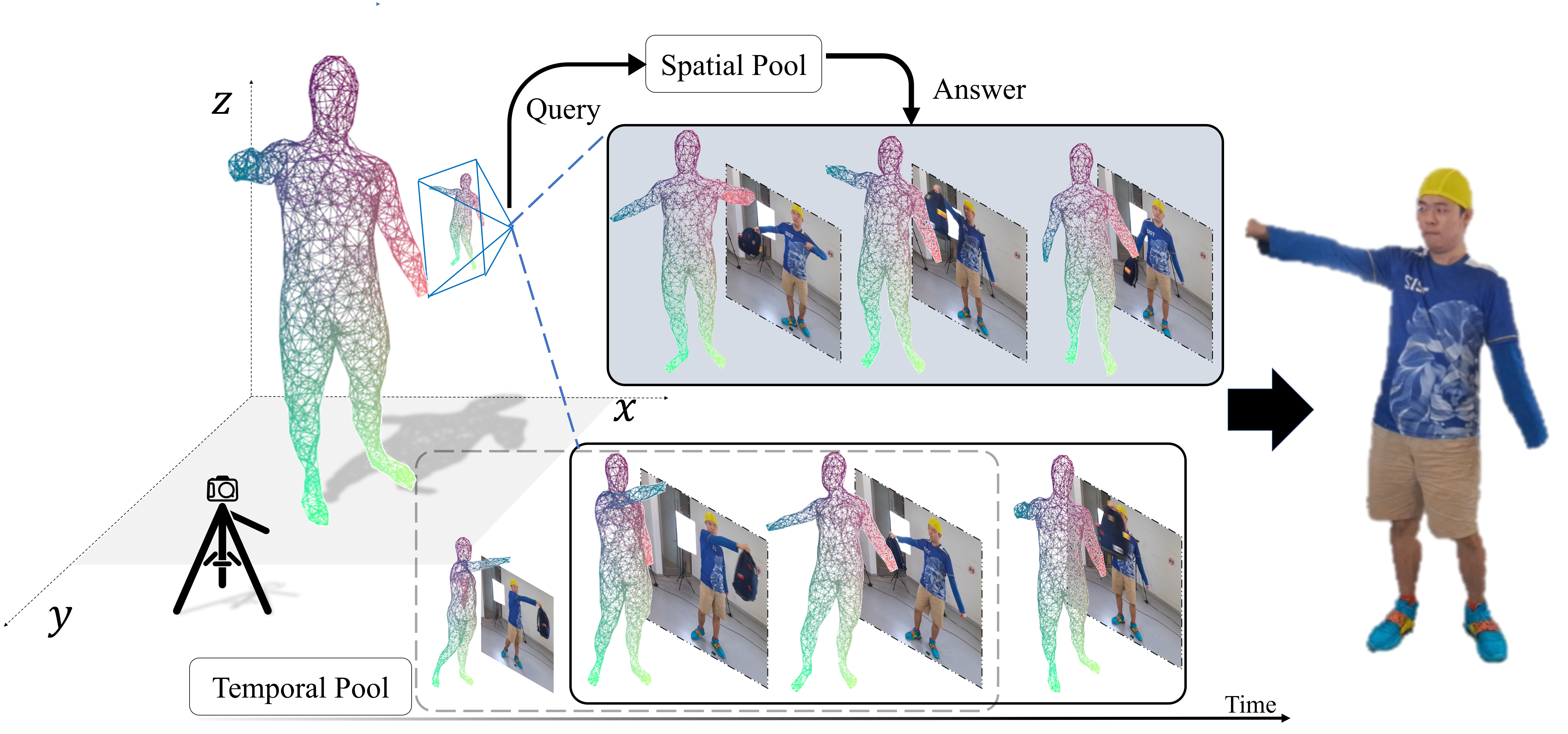}
	\end{center} 
	\vspace{-5mm}
	\caption{Illustration of our online refinement strategy.} 
	\label{fig:object_pipeline} 
	\vspace{-20pt}
\end{figure} 

\subsection{Neural Rendering Back-end}  \label{sec:Renderer}
To enable efficient photo-realistic neural rendering of the interaction scenes, our neural rendering back-end adopts a separated instant neural representation based on the on-the-fly key frame selection strategy.

\noindent{\bf Separated instant neural representation.}
We design the instant neural representation to reconstruct the human and object separately.
For object branch, given the RGBD image $I_t$ and $D_t$ with the camera pose $T^t$ as the training set, we leverage the original Instant-NGP~\cite{muller2022instant} to extract the 3D point $v^o_c$ features on the hash table $\bf H^o$ and then feed them into the geometry MLP $\bf E^o_g$ and color MLP $\bf E^o_c$ to acquire the density and color.

For dynamic human, in contrast to recent approaches~\cite{park2021nerfies, park2021hypernerf, tretschk2021nonrigid, pumarola2020d} which can't handle long sequences via pure MLP and human NeRFs~\cite{peng2021neural,weng_humannerf_2022_cvpr,zheng2022structured  } that heavily rely on SMPL~\cite{SMPL2015} which do not align well with the surface and easily cause artifacts,
we introduce a hybrid deformation module to efficiently leverage the motion priors. 
The explicit non-rigid warping and an implicit pose-conditioned deformation net jointly aggregate the corresponding point information in the canonical space. 

Specifically, given this human non-rigid motion $\{dq_i^t\}$, SMPL pose $\vec{\theta^t}$ and a sampling point $v^h_l$ at frame $t$, we construct the warping function 
to map $v^h_l$ back to the canonical space $v^h_t$.
We calculate the deformed ED nodes $ x^t_i = dq_i^t x_i$, and then the point $v^h_l$ in the influence radius $r$ of these nodes can be warped into canonical surface via neighboring ED nodes weight blending:
\begin{equation}
\begin{split}
\begin{aligned}
    v^h_t &= SE3( {DQB^{-1}}  (v_l^h)) v_l^h.
\end{aligned}
\end{split}
\label{equ4}
\end{equation}
To reduce the warping error and improve the rendering quality, we further integrate pose-conditioned deformation net here to correct the misalignment, where we concatenate the encoded $v^h_t$ via hash-encoding with the human pose $\vec{ \theta_t}$ and predict the residual displacement $\delta v^h$ through an MLP. 
Finally, we feed $v^h_c = v^h_t + \delta v^h$ into the canonical hash-encoding $\bf H^h$, geometry as well as color MLPs $ \bf E^h_g, E^h_c$.

\noindent{\bf On-the-fly Radiance Fields.} 
To ensure accurate tracking, hundreds of ED-nodes are maintained to query live points neighbors which is time-consuming and lead to bottlenecks.
 The time consumption is $O(n)$ even if we query a small number of neighbors for each sampled point, in which $n$ is the number of ED nodes. 
To enhance on-the-fly efficiency, we introduce a look-up-table-based fast search strategy here to speed up. Specifically, we only initialize the canonical KNN(k-nearest-neighbors) field in the beginning, whose resolution is $512^3$, and each voxel saves $s$ neighboring ED nodes index(4 in our setting). We then concatenate non-rigid motions in each frame to form a look-up table. 
At frame $t$, for a voxel with index $k$ and coordinates $v_k$ in the canonical, we warp it via Eqn.~\ref{equ2} to the live space and obtain its corresponding voxel index $f$. We save the canonical index $k$ in live voxel $f$. 
Afterward, for each sampling point, we can acquire the live space index $f$ and obtain the canonical index $k$. $k$ links the 4 neighbor ED nodes index.
Offsetting the index to frame $t$ on the look-up-table, we can acquire the corresponding motions and calculate the blending weight as well as warped point via $DQB$ in $O(1)$ manner. In addition, we are able to construct the live KNN field for each voxel in $O(1)$ time by utilizing custom CUDA kernels.

\noindent{\bf Online Key Frame Selection.} 
To achieve online performance and high quality rendering, we choose key frames to organize our neural rendering training dataset. Before choosing, we discard blurry RGB frames caused by fast motion based on the blurriness measure~\cite{crete2007blur}. 
Besides, we observe that naively selecting key-frames with fixed time intervals brings the time-related details but causes the non-evenly distribution of the captured regions.
Inspired by ~\cite{li2021posefusion,mustafa20164d}, we introduce a key frame selection scheme here to keep the diversity of motion distribution and complement visibility. 
Specifically, we formulate the visibility map for each ED node $x^t_i = (x',y',z')$ in frame $t$ as follows: 
\begin{equation}
\label{eq:alpha}
s^{t}_i =\left\{
 \begin{array}{lcl}
 1,\quad if\quad | z' - D^t(\pi (x^t_i)) | < \epsilon \\
 0,\quad othersize 
 \end{array},
 \right.
\end{equation}
where $\pi(\cdot)$ denotes the projection matrix, $D^t(\cdot)$ represents the depth value of the corresponding pixel at frame $t$, $\epsilon$ is the visibility degree (0.01 in our setting).
we continue to define the similarity for two frames: 
\begin{equation}
\begin{aligned}
E_h(t_1,t_2) =  & \vec{\beta_{\mathrm{pose}}} | \vec{\theta_{t_1}} - \vec{\theta_{t_2}} |^2 + \beta_{\mathrm{vis}} \sum_{i}{s_i^{t_1} \oplus s_i^{t_2}}   + \\ &\beta_{\mathrm{h}}| t_1 - t_2 |^2,
\end{aligned}
\label{equ7}
\end{equation}
where $\oplus$ is the xor  operation and  $t_1,t_2$ are frame indexes.

For an object with the pose $T^t$ which includes rotation $\mathbf{R}^t$ and translation $\vec{{d}^t}$, we define the similarity as follows:
\begin{equation}
\begin{aligned}
E_o(t_1,t_2) = \beta_{\mathrm{d}} \| \vec{{d}^{t_1}} - \vec{{d}^{t_2}}\|^2_2 + \beta_{\mathrm{o}}| t_1 - t_2|^2,
\end{aligned}
\label{equ7}
\end{equation}

\begin{figure*}[htbp] 
	\begin{center} 
		\includegraphics[width=1.0\linewidth]{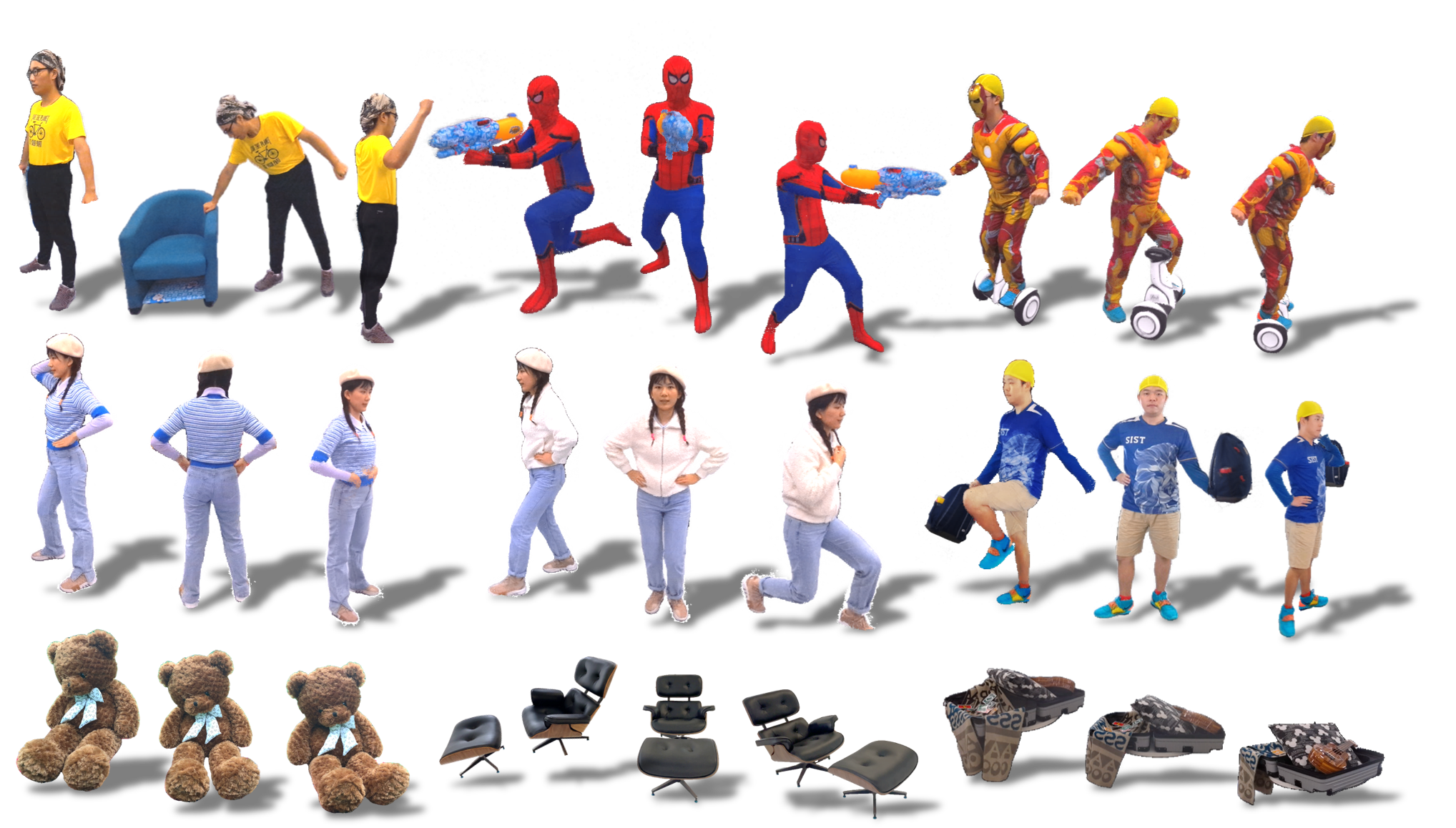} 
	\end{center} 
	\vspace{-10pt}
	\caption{The rendering results of Instant-NVR on various interaction sequences, including ``driving a balance car'',``shaking a bag'', and ``playing a water gun''.} 
	\label{fig:fig_all}
	\vspace{-10pt}
\end{figure*}

Furthermore, we define $\gamma$ to determine the diversity of the spatial pool.
At the start time, the pool is empty and imported the first frame. 
Once the similarity of each two among the latest frame received from tracking and frame(s) in the pool is greater than $\gamma$, we push this frame to the pool. 
When the pool capacity reaches its peak (100 in our setting), we will 
continually update the pool using the new frame by removing the frame with the biggest similarity. In this manner, our spatial pool constantly updates in all frames.

To achieve photo-realistic novel-view synthesis, our Instant-NVR further refines the rendering view via training short iterations on the carefully selected frames in a spatial-temporal pool. The spatial-temporal pool includes $m$ frames from the spatial pool, which have the most similarity with the rendering view and $m$ latest frames received from the front-end. 
Our selection strategy ensures high-quality rendering without losing temporal detail.

\subsection{Implementation Details} \label{sec:detail}

To train the dynamic NeRF under human-object interactions, we first apply the semantic segmentation MIVOS~\cite{cheng2021modular} to decouple the scene and obtain the human and object masks separately. 
To assemble human and object in a novel view,
we additionally render the depth maps and then combine the RGB images according to the depth occlusion.
We implement our entire pipeline on GPU based on the Instant-NGP~\cite{muller2022instant} using two Nvidia GeForce RTX3090 GPU. One GPU for the tracking front-end(14 GB memory consumption) and another for the neural rendering(15 GB for human branch and 4 GB for object).
For deformation net, the input is the 32-dim hash feature and the 72-dim pose. The hidden layer is 4 and hidden dimension is 128. 
For key frame selection, we use the bigger weight for the joint rotation of the torso to domain the human pose diversity, specifically, for $\vec{\beta_{\mathrm{pose}}}$, we set each torso weight $\beta_{pose(torso)} = 0.1$ and each limbs weight as 0.02.
Besides, we use the following empirically determined parameters: $\beta_{\mathrm{vis}}=0.01, \beta_{\mathrm{h}}= 0.02, \beta_{\mathrm{d}}=1.0,  \beta_{\mathrm{o}}=0.02, \gamma = 2.5, \lambda_{\mathrm{data}}=1.0, \lambda_{\mathrm{bind}}= 1.0, \lambda_{\mathrm{reg}}=4.0, \lambda_{\mathrm{prior}}=0.01,  \lambda_{\mathrm{pose}}=0.02, \lambda_{\mathrm{inter}}=1.0$. 
For efficiency, we choose $m=10$ in the spatial-temporal pool. We use the photometric loss and depth loss to supervise human NeRF and object NeRF separately as follows:
\begin{equation}
\begin{aligned}
\mathcal{L}_{\text {color }} &=\sum_{\mathbf{r} \in    \mathcal{R}}\|M(\mathbf{r})(\hat{\mathbf{C}}(\mathbf{r})-\mathbf{C}(\mathbf{r}))\|_2, \\
\mathcal{L}_{\text {depth }} &=\sum_{\mathbf{r} \in \mathcal{R}}\|M(\mathbf{r})(\hat{\mathbf{D}}(\mathbf{r})-\mathbf{D}(\mathbf{r}))\|_1
\end{aligned}
\end{equation}
where $M(r)$ is human or object mask.

    \begin{figure*}[htbp] 
	\begin{center} 
		\includegraphics[width=0.95\linewidth]{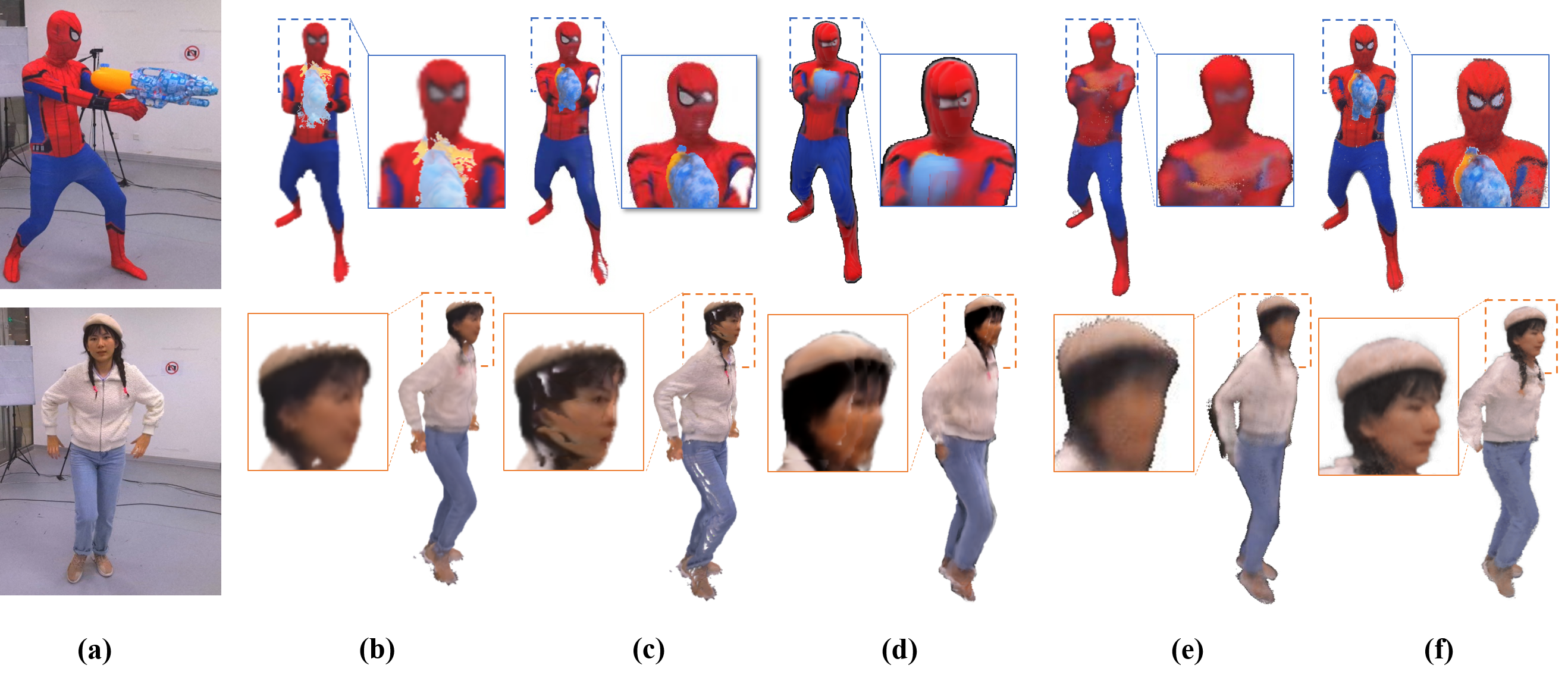} 
	\end{center} 
	\vspace{-20pt}
	\caption{Qualitative comparison with fusion-based methods and NeRF-based methods. (a) Reference view. (b) RobustFusion~\cite{su2022robustfusionPlus} (c) NeuralHOFusion~\cite{jiang2022neuralhofusion} (d) NeuralBody~\cite{peng2021neural} (e) HumanNerf~\cite{zhao2022humannerf} (f) Ours.}
	\label{fig:fig_comp_1}
	\vspace{-20pt} 
\end{figure*} 

\section{Experimental Results} 
In this section, we compare the state-of-the-art methods and evaluate  Instant-NVR on various challenging human-object interaction scenarios. Besides, various rendering results of Instant-NVR are shown in Fig.~\ref{fig:fig_all}, such as driving a balance car, shaking a bag and playing with a water gun. Please also kindly refer to our video.

\subsection{Comparison} 
We compare Instant-NVR against the fusion-based methods RobustFusion~\cite{su2022robustfusionPlus}, 
NeuralHOFusion~\cite{jiang2022neuralhofusion} and NeRF-based methods NeuralBody~\cite{peng2021neural}, HumanNerf~\cite{zhao2022humannerf}, both in efficiency and rendering quality.
For comparison with fusion-based methods,
as illustrated in Fig.~\ref{fig:fig_comp_1} (b), RobustFusion~\cite{su2022robustfusionPlus} generates blurry appearance results, which are restricted by the limited geometry resolution.
For a fair comparison, NeuralHOFusion~\cite{jiang2022neuralhofusion} is modified to a single view setting and suffers from artifacts as shown in Fig.~\ref{fig:fig_comp_1} (c). 
For NeRF-based methods, we employ the RGBD input to estimate the SMPL~\cite{SMPL2015} as their prior and adopt RGBD loss terms.
Both NeuralBody~\cite{peng2021neural} and HumanNerf~\cite{zhao2022humannerf} give erroneous and blurry rendering results in the monocular setting (Fig.~\ref{fig:fig_comp_1} (d-e)), which rely heavily on SMPL~\cite{SMPL2015} and can not handle human-object interactions. In addition, training in these methods is time-consuming and novel view synthesis remains slow.
In contrast, our Instant-NVR achieves more detailed and photo-realistic rendering results under human-object interactions, as shown in Fig.~\ref{fig:fig_comp_1} (f).
The quantitative results in Tab.~\ref{table:Comparison_texture} also demonstrate that our approach can achieve consistently better rendering quality and achieve efficient training as well as rendering speed to support on-the-fly performance. Note that both NeuralBody~\cite{peng2021neural} and HumanNerf~\cite{zhao2022humannerf} take several hours to train, while training for our method is online.

\begin{table}[!htbp]
	\begin{center}
		\centering
		\caption{Comparison against fusion and NeRF-based methods.}
		\vspace{-10pt}
		\label{table:Comparison_texture}
		\resizebox{0.45\textwidth}{!}{
			\begin{tabular}{l|cccc}
				\hline
				Method     & PSNR$\uparrow$  & SSIM $\uparrow$ & Rendering Time$\downarrow$  \\
				\hline
				RobustFusion ~\cite{su2022robustfusionPlus}\qquad\qquad & 20.59           & 0.935          & 0.123s    \\ %
				NeuralHOFusion  ~\cite{jiang2022neuralhofusion}       & 21.09         &0.942          & ~0.151s  \\  %
				NeuralBody  ~\cite{peng2021neural} &      19.71  & 0.928 & ~2.420s \\
				HumanNerf  ~\cite{zhao2022humannerf} &      18.68  & 0.892 & 5.103s \\
				Ours        & \textbf{27.81}   & \textbf{0.976} & \textbf{0.023s} \\  %
				\hline
			\end{tabular}
		}
		\vspace{-20pt}
	\end{center}
\end{table}

\subsection{Evaluation} \label{sec:abla} 

\begin{figure}[t] 
	\begin{center} 
		\includegraphics[width=1.0\linewidth]{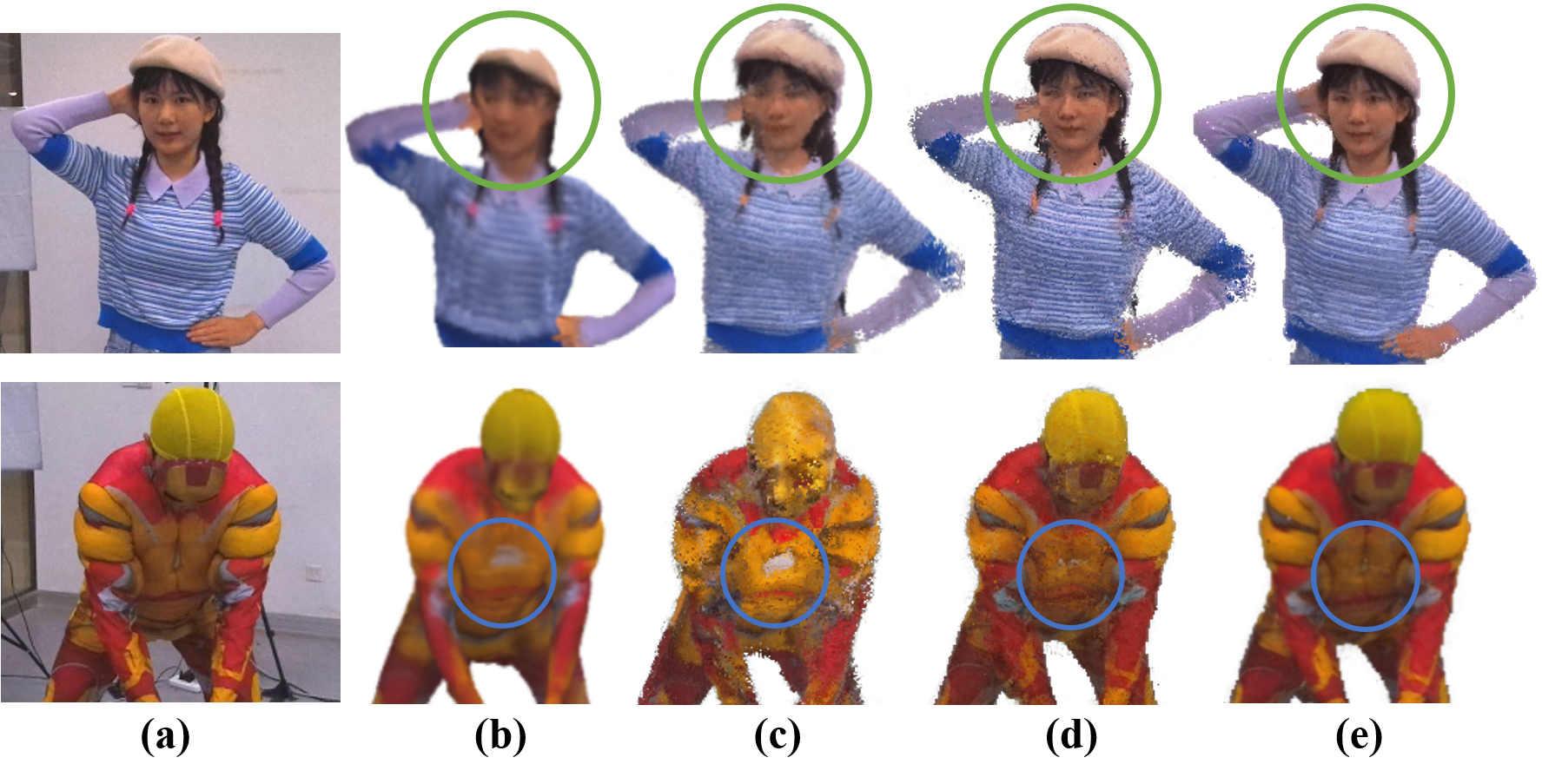} 
	\end{center} 
	\vspace{-10pt}
	\caption{Quantitative evaluation of Online Human Rendering. (a) Input image; (b) Per-vertex texture; (c) Key frame selection using fixed interval; (c) Key frame selection w/o refinement; (d) Key frame selection w refinement.} 
	\label{fig:eval_tex_human} 
	\vspace{-10pt}
\end{figure}

\myparagraph{Online Human rendering.}
As shown in Fig.~\ref{fig:eval_tex_human} (b), per-vertex texture extracted from the fused albedo volume~\cite{guo2017real} is blurry.
Naively selecting key-frames with fixed time intervals generates noising rendering results in Fig.~\ref{fig:eval_tex_human} (c) due to the non-evenly distribution of the captured regions.
In contrast, our online key frame selection strategy based on the diversity of motion distribution and complement visibility can achieve much clearer rendering results, as shown in  Fig.~\ref{fig:eval_tex_human} (d).
To boost the rendering quality, the further refinement scheme can help us to achieve more photo-realistic rendering results, as shown in Fig.~\ref{fig:eval_tex_human} (e).
As for quantitative analysis, we evaluate the rendering quality in Tab.~\ref{table:eval_tex_human}, which highlights the contributions of each component.

\begin{table}[t]
	\begin{center}
		\centering
		\caption{Quantitative evaluation of Online Human Rendering.}
		\vspace{-10pt}
		\label{table:eval_tex_human}
		\resizebox{0.45\textwidth}{!}{
			\begin{tabular}{l|cccc}
				\hline
				Method     & PSNR $\uparrow$& SSIM $\uparrow$ & MAE $\downarrow$\\
				\hline
				Per-vertex texture\qquad\qquad & 19.710     & 0.902         & 3.176    \\
				Key frame selection using fixed interval      & 23.071  & 0.922         & 1.571  \\
				Key frame selection w/o refinement  &      25.768  & 0.949 & 1.229 \\
				Key frame selection w refinement  &     \textbf{ 28.255}  & \textbf{0.972} & \textbf{0.534} \\
				\hline
			\end{tabular}
		}
		\vspace{-10pt}
	\end{center}
\end{table}

\begin{figure}[t] 
	\begin{center} 
		\includegraphics[width=1\linewidth]{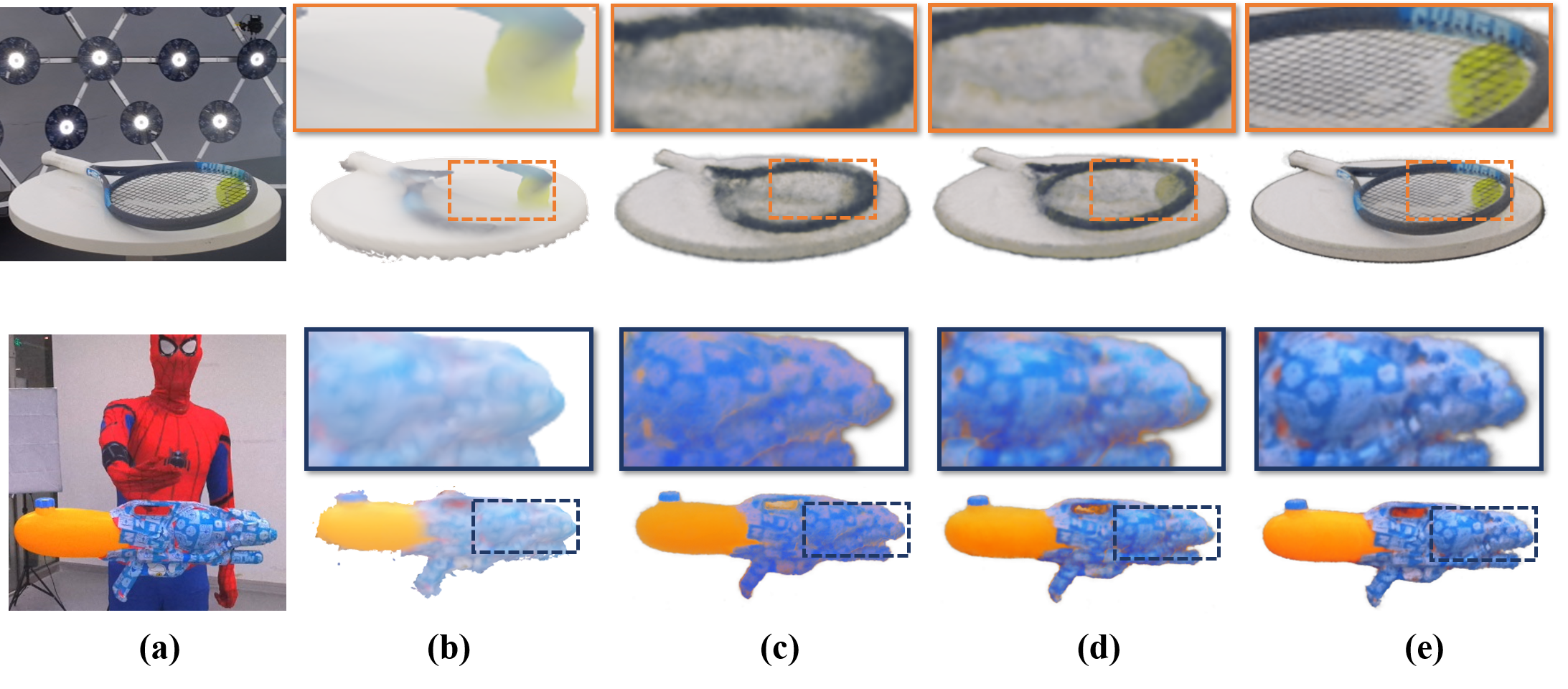} 
	\end{center} 
	\vspace{-10pt}
	\caption{Quantitative evaluation of Online Object Rendering. (a) Input image; (b) Per-vertex texture; (c) Key frame selection using fixed interval; (c) Key frame selection w/o refinement ; (d) Key frame selection w refinement.} 
	\label{fig:eval_tex_obj} 
	\vspace{-10pt}
\end{figure}

\myparagraph{Online Object Rendering.}
As for the evaluation of online object rendering in Fig.~\ref{fig:eval_tex_obj}, we can observe that per-vertex texture failed to generate high-quality appearance which is restricted by the limited geometry resolution. Moreover, naively selecting key-frames with fixed time interval brings the noises. Fig.~\ref{fig:eval_tex_obj} (d) shows that applying our key frame selection strategy without refinement is still unclear.
In contrast, we can achieve the best rendering results with our full training pipeline.
Moreover, the quantitative evaluation is as demonstrated in Tab.~\ref{table:eval_tex_obj}, in which our full pipeline with the online key frame selection and rendering refinement achieves the highest accuracy.

\begin{table}[t]
	\begin{center}
		\centering
		\caption{Quantitative evaluation of Online Object Rendering.}
		\vspace{-10pt}
		\label{table:eval_tex_obj}
		\resizebox{0.45\textwidth}{!}{
			\begin{tabular}{l|cccc}
				\hline
				Method     & PSNR $\uparrow$ & SSIM $\uparrow$ & MAE $\downarrow$\\
				\hline
				Per-vertex texture\qquad\qquad & 21.253    & 0.944         & 4.431    \\
				Key frame selection using fixed interval  & 25.248         & 0.954         & 1.043  \\
				Key frame selection w/o refinement   &      26.747  & 0.965 & 0.931 \\
				Key frame selection w refinement   &      \textbf{28.826}  & \textbf{0.977} & \textbf{0.615} \\
				\hline
			\end{tabular}
		}
		\vspace{-10pt}
	\end{center}
\end{table}

\myparagraph{Run-time Evaluation.}
 In Tab.~\ref{table:breakout}, we list the run-time of each step in our pipeline, including both the tracking front-end and the neural rendering back-end. 
 For tracking front-end, the rigid tracking of the object takes 40ms while the human non-rigid tracking takes 62ms. Besides, the deformation net costs 5ms.
 For rendering back-end, training without fast search strategy takes 205.53ms while using our fast search scheme, the training time reduces to 17.95ms. Besides, we use 15.38ms for the rendering process.

\begin{table}[t]\tiny
	\begin{center}
		\centering
		\caption{Quantitative evaluation of Run-time}
		\vspace{-10pt}
		\label{table:breakout}
		\resizebox{0.4\textwidth}{!}{
			\begin{tabular}{l|cccc}
				\hline
				     Procedure  &  Time\\
				\hline
			    	 rigid tracking   \quad\quad\quad \quad\quad\quad\quad    &\quad\quad\quad 40ms   \quad\quad\quad\quad\\
					 non-rigid-tracking                 & 62ms     \\
			    	 deformation net                     & 5ms  \\
				\hline
					 training w/o fast search            &      205.53ms \\
			    	 training w fast search             & 17.95ms  \\
					 rendering                          &      23.38ms \\
				\hline
			\end{tabular}
		}
		\vspace{-10pt}
	\end{center}
\end{table}

\begin{figure}[tbp] 
	\begin{center} 
		\includegraphics[width=1\linewidth]{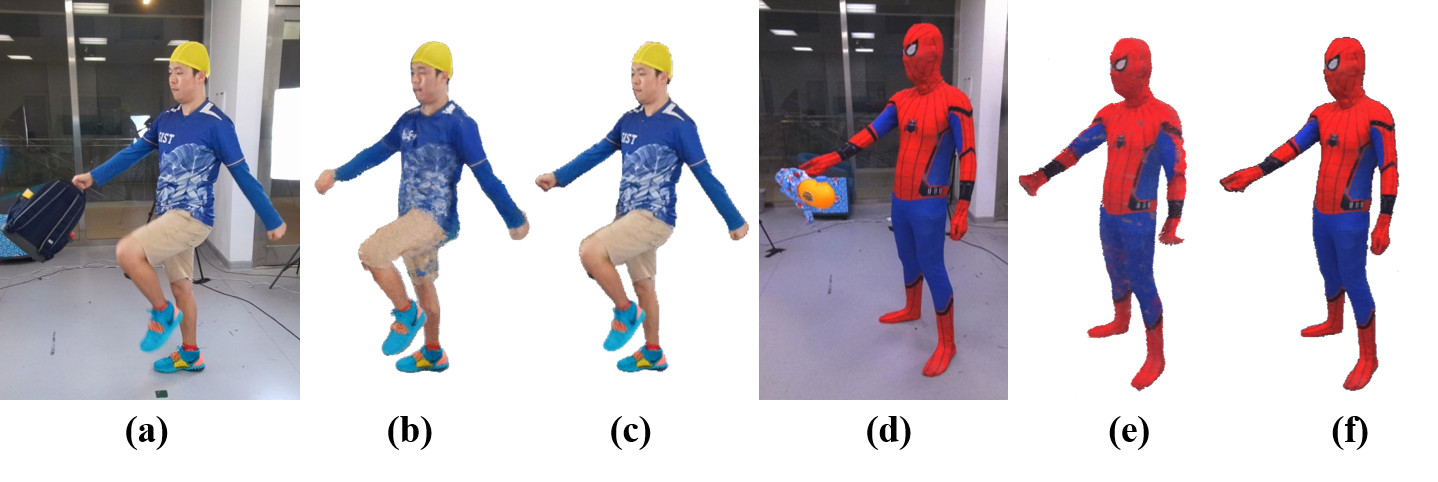} 
	\end{center} 
	\vspace{-20pt}
	\caption{Evaluation of the hybrid deformation module. (a), (d) are the reference views. (b), (e) are the results with only deform block. (c),(f) use our deform block with the aid of deform net.} 
	\label{fig:eval_deform} 
	\vspace{-4mm}
\end{figure}

\myparagraph{Hybrid Deformation Module Evaluation.}
We conduct further evaluation of our hybrid deformation module to demonstrate its advantages. As shown in Fig.~\ref{fig:eval_deform} (b)(e), employing only the explicit deform block results in misalignment between the ground truth and warped space, leading to blurry images and erroneous silhouettes. Conversely, by utilizing the deform block with the aid of implicit deform net to learn the residual displacement in Fig.~\ref{fig:eval_deform} (c)(f), the rendering results outcome exhibit superior alignment and significantly enhance texture.

\subsection{Limitation}
As the first instant neural rendering system from an RGBD sensor that performs real-time and photo-realistic novel-view synthesis under human-object interactions, the proposed Instant-NVR still has some limitations. 
First, although we adopt the hybrid deformation module to efficiently utilize the non-rigid motion priors since our method is in monocular RGB-D camera setting, non-rigid fusion fails when facing the fast movement and leads to inaccurate priors which affect the on-the-fly rendering.
Due to limited resolution and inherent noise of the depth input, our method cannot reconstruct the extremely fine details of the performer,
such as the fingers. Data-driven techniques on different human
parts will be critical for such problem.
Besides, Instant-NVR is committed to rendering photo-realistic results on-the-fly. Therefore, we choose the density field as geometry representation, analogous to Instant-NGP~\cite{muller2022instant}. It is promising to integrate other SDF representations\cite{wang2021neus,Cai2022NDR}, which can generate a more delicate geometry.
Furthermore, to ensure efficient transmission between tracking front-end and rendering back-end, we discard the volumetric explicit geometry priors produced by the tracking step. It is an interesting direction to explore more complementary between tracking and rendering.

\section {Conclusion} 
We have presented a practical neural tracking and rendering approach for human-object interaction scenes using a single RGBD camera. By bridging traditional non-rigid tracking with recent instant radiance field techniques, our system achieves a photo-realistic free-viewing experience for human-object scenes on the fly.
Our non-rigid tracking robustly provides sufficient motion priors for both the performer and the object.
Our separated instant neural representation with hybrid deformation and efficient motion-prior searching enables the on-the-fly reconstruction of both the dynamic and static radiance fields.
Our online key frame selection with a rendering-aware refinement strategy further provides a more vivid and detailed novel-view synthesis for our online setting.
Our experimental results demonstrate the effectiveness of Instant-NVR for the instant generation of dynamic radiance fields and photo-realistic novel view synthesis of human-object interactions in real time.
We believe that our approach is a critical step to virtual but realistic teleport human-object interactions, with many potential applications like consumer-level telepresence in VR/AR. \\
\noindent{\bf Acknowledgements.} This work was supported by Shanghai YangFan Program (21YF1429500), Shanghai Local college capacity building program (22010502800). We also acknowledge support from Shanghai Frontiers Science Center of Human-centered Artificial Intelligence (ShangHAI).

{\small
\bibliographystyle{ieee_fullname}
\bibliography{reference}

\begin{thebibliography}{10}\itemsep=-1pt

\bibitem{aliev2019neural}
Kara-Ali Aliev, Artem Sevastopolsky, Maria Kolos, Dmitry Ulyanov, and Victor
  Lempitsky.
\newblock Neural point-based graphics.
\newblock In {\em European Conference on Computer Vision}, pages 696--712.
  Springer, 2020.

\bibitem{bhatnagar22behave}
Bharat~Lal Bhatnagar, Xianghui Xie, Ilya Petrov, Cristian Sminchisescu,
  Christian Theobalt, and Gerard Pons-Moll.
\newblock Behave: Dataset and method for tracking human object interactions.
\newblock In {\em {IEEE} Conference on Computer Vision and Pattern Recognition
  (CVPR)}. {IEEE}, jun 2022.

\bibitem{keepitSMPL}
Federica Bogo, Angjoo Kanazawa, Christoph Lassner, Peter Gehler, Javier Romero,
  and Michael~J. Black.
\newblock Keep it smpl: Automatic estimation of 3d human pose and shape from a
  single image.
\newblock In Bastian Leibe, Jiri Matas, Nicu Sebe, and Max Welling, editors,
  {\em Computer Vision -- ECCV 2016}, pages 561--578, Cham, 2016. Springer
  International Publishing.

\bibitem{bozic2021neural}
Aljaz Bozic, Pablo Palafox, Michael Zollhofer, Justus Thies, Angela Dai, and
  Matthias Nie{\ss}ner.
\newblock Neural deformation graphs for globally-consistent non-rigid
  reconstruction.
\newblock In {\em Proceedings of the IEEE/CVF Conference on Computer Vision and
  Pattern Recognition}, pages 1450--1459, 2021.

\bibitem{bozic2020deepdeform}
Aljaz Bozic, Michael Zollhofer, Christian Theobalt, and Matthias Nie{\ss}ner.
\newblock Deepdeform: Learning non-rigid rgb-d reconstruction with
  semi-supervised data.
\newblock In {\em Proceedings of the IEEE/CVF Conference on Computer Vision and
  Pattern Recognition}, pages 7002--7012, 2020.

\bibitem{bradley2008markerless}
Derek Bradley, Tiberiu Popa, Alla Sheffer, Wolfgang Heidrich, and Tamy
  Boubekeur.
\newblock Markerless garment capture.
\newblock In {\em ACM SIGGRAPH 2008 papers}, pages 1--9. 2008.

\bibitem{Cai2022NDR}
Hongrui Cai, Wanquan Feng, Xuetao Feng, Yan Wang, and Juyong Zhang.
\newblock Neural surface reconstruction of dynamic scenes with monocular rgb-d
  camera.
\newblock In {\em Thirty-sixth Conference on Neural Information Processing
  Systems (NeurIPS)}, 2022.

\bibitem{cheng2021modular}
Ho~Kei Cheng, Yu-Wing Tai, and Chi-Keung Tang.
\newblock Modular interactive video object segmentation: Interaction-to-mask,
  propagation and difference-aware fusion.
\newblock In {\em Proceedings of the IEEE/CVF Conference on Computer Vision and
  Pattern Recognition}, pages 5559--5568, 2021.

\bibitem{collet2015high}
Alvaro Collet, Ming Chuang, Pat Sweeney, Don Gillett, Dennis Evseev, David
  Calabrese, Hugues Hoppe, Adam Kirk, and Steve Sullivan.
\newblock High-quality streamable free-viewpoint video.
\newblock {\em ACM Transactions on Graphics (TOG)}, 34(4):69, 2015.

\bibitem{crete2007blur}
Frederique Crete, Thierry Dolmiere, Patricia Ladret, and Marina Nicolas.
\newblock The blur effect: perception and estimation with a new no-reference
  perceptual blur metric.
\newblock In {\em Human vision and electronic imaging XII}, volume 6492, pages
  196--206. SPIE, 2007.

\bibitem{motion2fusion}
Mingsong Dou, Philip Davidson, Sean~Ryan Fanello, Sameh Khamis, Adarsh Kowdle,
  Christoph Rhemann, Vladimir Tankovich, and Shahram Izadi.
\newblock Motion2fusion: Real-time volumetric performance capture.
\newblock {\em ACM Trans. Graph.}, 36(6):246:1--246:16, Nov. 2017.

\bibitem{dou2016fusion4d}
Mingsong Dou, Sameh Khamis, Yury Degtyarev, Philip Davidson, Sean~Ryan Fanello,
  Adarsh Kowdle, Sergio~Orts Escolano, Christoph Rhemann, David Kim, Jonathan
  Taylor, et~al.
\newblock Fusion4d: Real-time performance capture of challenging scenes.
\newblock {\em ACM Transactions on Graphics (ToG)}, 35(4):1--13, 2016.

\bibitem{guo2019relightables}
Kaiwen Guo, Peter Lincoln, Philip Davidson, Jay Busch, Xueming Yu, Matt Whalen,
  Geoff Harvey, Sergio Orts-Escolano, Rohit Pandey, Jason Dourgarian, et~al.
\newblock The relightables: Volumetric performance capture of humans with
  realistic relighting.
\newblock {\em ACM Transactions on Graphics (TOG)}, 38(6):1--19, 2019.

\bibitem{guo2017real}
Kaiwen Guo, Feng Xu, Tao Yu, Xiaoyang Liu, Qionghai Dai, and Yebin Liu.
\newblock Real-time geometry, albedo and motion reconstruction using a single
  rgbd camera.
\newblock {\em ACM Transactions on Graphics (TOG)}, 2017.

\bibitem{haresh2022articulated}
Sanjay Haresh, Xiaohao Sun, Hanxiao Jiang, Angel~X Chang, and Manolis Savva.
\newblock Articulated 3d human-object interactions from rgb videos: An
  empirical analysis of approaches and challenges.
\newblock In {\em 2022 International Conference on 3D Vision (3DV)}. IEEE,
  2022.

\bibitem{innmann2016volumedeform}
Matthias Innmann, Michael Zollh{\"o}fer, Matthias Nie{\ss}ner, Christian
  Theobalt, and Marc Stamminger.
\newblock Volumedeform: Real-time volumetric non-rigid reconstruction.
\newblock In {\em European Conference on Computer Vision}, pages 362--379.
  Springer, 2016.

\bibitem{jiang2022neuralhofusion}
Yuheng Jiang, Suyi Jiang, Guoxing Sun, Zhuo Su, Kaiwen Guo, Minye Wu, Jingyi
  Yu, and Lan Xu.
\newblock Neuralhofusion: Neural volumetric rendering under human-object
  interactions.
\newblock In {\em Proceedings of the IEEE/CVF Conference on Computer Vision and
  Pattern Recognition}, pages 6155--6165, 2022.

\bibitem{TotalCapture}
Hanbyul Joo, Tomas Simon, and Yaser Sheikh.
\newblock Total capture: A 3d deformation model for tracking faces, hands, and
  bodies.
\newblock In {\em The IEEE Conference on Computer Vision and Pattern
  Recognition (CVPR)}, June 2018.

\bibitem{li2009robust}
Hao Li, Bart Adams, Leonidas~J Guibas, and Mark Pauly.
\newblock Robust single-view geometry and motion reconstruction.
\newblock {\em ACM Transactions on Graphics (ToG)}, 28(5):1--10, 2009.

\bibitem{li2022tava}
Ruilong Li, Julian Tanke, Minh Vo, Michael Zollhofer, Jurgen Gall, Angjoo
  Kanazawa, and Christoph Lassner.
\newblock Tava: Template-free animatable volumetric actors.
\newblock 2022.

\bibitem{li2021neural}
Tianye Li, Mira Slavcheva, Michael Zollhoefer, Simon Green, Christoph Lassner,
  Changil Kim, Tanner Schmidt, Steven Lovegrove, Michael Goesele, and Zhaoyang
  Lv.
\newblock Neural 3d video synthesis, 2021.

\bibitem{li2021posefusion}
Zhe Li, Tao Yu, Zerong Zheng, Kaiwen Guo, and Yebin Liu.
\newblock Posefusion: Pose-guided selective fusion for single-view human
  volumetric capture.
\newblock In {\em IEEE Conference on Computer Vision and Pattern Recognition},
  June 2021.

\bibitem{lin2022occlusionfusion}
Wenbin Lin, Chengwei Zheng, Jun-Hai Yong, and Feng Xu.
\newblock Occlusionfusion: Occlusion-aware motion estimation for real-time
  dynamic 3d reconstruction.
\newblock In {\em Proceedings of the IEEE/CVF Conference on Computer Vision and
  Pattern Recognition}, pages 1736--1745, 2022.

\bibitem{liu2022devrf}
Jia-Wei Liu, Yan-Pei Cao, Weijia Mao, Wenqiao Zhang, David~Junhao Zhang, Jussi
  Keppo, Ying Shan, Xiaohu Qie, and Mike~Zheng Shou.
\newblock Devrf: Fast deformable voxel radiance fields for dynamic scenes.
\newblock {\em arXiv preprint arXiv:2205.15723}, 2022.

\bibitem{liu2019neural}
Lingjie Liu, Weipeng Xu, Michael Zollhoefer, Hyeongwoo Kim, Florian Bernard,
  Marc Habermann, Wenping Wang, and Christian Theobalt.
\newblock Neural rendering and reenactment of human actor videos.
\newblock {\em ACM Transactions on Graphics (TOG)}, 38(5):1--14, 2019.

\bibitem{lombardi2019neural}
Stephen Lombardi, Tomas Simon, Jason Saragih, Gabriel Schwartz, Andreas
  Lehrmann, and Yaser Sheikh.
\newblock Neural volumes: Learning dynamic renderable volumes from images.
\newblock {\em ACM Trans. Graph.}, 38(4):65:1--65:14, July 2019.

\bibitem{SMPL2015}
Matthew Loper, Naureen Mahmood, Javier Romero, Gerard Pons-Moll, and Michael~J.
  Black.
\newblock Smpl: A skinned multi-person linear model.
\newblock {\em ACM Trans. Graph.}, 34(6):248:1--248:16, Oct. 2015.

\bibitem{luo2022artemis}
Haimin Luo, Teng Xu, Yuheng Jiang, Chenglin Zhou, Qiwei Qiu, Yingliang Zhang,
  Wei Yang, Lan Xu, and Jingyi Yu.
\newblock Artemis: Articulated neural pets with appearance and motion
  synthesis.
\newblock {\em ACM Trans. Graph.}, 41(4), jul 2022.

\bibitem{nerf}
Ben Mildenhall, Pratul~P. Srinivasan, Matthew Tancik, Jonathan~T. Barron, Ravi
  Ramamoorthi, and Ren Ng.
\newblock Nerf: Representing scenes as neural radiance fields for view
  synthesis.
\newblock In Andrea Vedaldi, Horst Bischof, Thomas Brox, and Jan-Michael Frahm,
  editors, {\em Computer Vision -- ECCV 2020}, pages 405--421, Cham, 2020.
  Springer International Publishing.

\bibitem{muller2022instant}
Thomas M\"uller, Alex Evans, Christoph Schied, and Alexander Keller.
\newblock Instant neural graphics primitives with a multiresolution hash
  encoding.
\newblock {\em ACM Trans. Graph.}, 41(4):102:1--102:15, July 2022.

\bibitem{mustafa20164d}
Armin Mustafa, Hansung Kim, and Adrian Hilton.
\newblock 4d match trees for non-rigid surface alignment.
\newblock In {\em European Conference on Computer Vision}, pages 213--229.
  Springer, 2016.

\bibitem{newcombe2015dynamicfusion}
Richard~A Newcombe, Dieter Fox, and Steven~M Seitz.
\newblock Dynamicfusion: Reconstruction and tracking of non-rigid scenes in
  real-time.
\newblock In {\em Proceedings of the IEEE conference on computer vision and
  pattern recognition}, pages 343--352, 2015.

\bibitem{oechsle2021unisurf}
Michael Oechsle, Songyou Peng, and Andreas Geiger.
\newblock Unisurf: Unifying neural implicit surfaces and radiance fields for
  multi-view reconstruction.
\newblock In {\em Proceedings of the IEEE/CVF International Conference on
  Computer Vision}, pages 5589--5599, 2021.

\bibitem{DeepSDF}
Jeong~Joon Park, Peter Florence, Julian Straub, Richard Newcombe, and Steven
  Lovegrove.
\newblock Deepsdf: Learning continuous signed distance functions for shape
  representation.
\newblock In {\em Proceedings of the IEEE/CVF Conference on Computer Vision and
  Pattern Recognition (CVPR)}, June 2019.

\bibitem{park2021nerfies}
Keunhong Park, Utkarsh Sinha, Jonathan~T Barron, Sofien Bouaziz, Dan~B Goldman,
  Steven~M Seitz, and Ricardo Martin-Brualla.
\newblock Nerfies: Deformable neural radiance fields.
\newblock In {\em Proceedings of the IEEE/CVF International Conference on
  Computer Vision}, pages 5865--5874, 2021.

\bibitem{park2021hypernerf}
Keunhong Park, Utkarsh Sinha, Peter Hedman, Jonathan~T. Barron, Sofien Bouaziz,
  Dan~B Goldman, Ricardo Martin-Brualla, and Steven~M. Seitz.
\newblock Hypernerf: A higher-dimensional representation for topologically
  varying neural radiance fields.
\newblock {\em ACM Trans. Graph.}, 40(6), dec 2021.

\bibitem{peng2021neural}
Sida Peng, Yuanqing Zhang, Yinghao Xu, Qianqian Wang, Qing Shuai, Hujun Bao,
  and Xiaowei Zhou.
\newblock Neural body: Implicit neural representations with structured latent
  codes for novel view synthesis of dynamic humans.
\newblock In {\em Proceedings of the IEEE/CVF Conference on Computer Vision and
  Pattern Recognition}, pages 9054--9063, 2021.

\bibitem{pumarola2020d}
Albert Pumarola, Enric Corona, Gerard Pons-Moll, and Francesc Moreno-Noguer.
\newblock D-nerf: Neural radiance fields for dynamic scenes.
\newblock In {\em Proceedings of the IEEE/CVF Conference on Computer Vision and
  Pattern Recognition}, pages 10318--10327, 2021.

\bibitem{yu_and_fridovichkeil2021plenoxels}
{Sara Fridovich-Keil and Alex Yu}, Matthew Tancik, Qinhong Chen, Benjamin
  Recht, and Angjoo Kanazawa.
\newblock Plenoxels: Radiance fields without neural networks.
\newblock In {\em CVPR}, 2022.

\bibitem{shao2022doublefield}
Ruizhi Shao, Hongwen Zhang, He Zhang, Mingjia Chen, Yanpei Cao, Tao Yu, and
  Yebin Liu.
\newblock Doublefield: Bridging the neural surface and radiance fields for
  high-fidelity human reconstruction and rendering.
\newblock In {\em CVPR}, 2022.

\bibitem{KillingFusion2017cvpr}
Miroslava Slavcheva, Maximilian Baust, Daniel Cremers, and Slobodan Ilic.
\newblock Killingfusion: Non-rigid 3d reconstruction without correspondences.
\newblock In {\em Proceedings of the IEEE Conference on Computer Vision and
  Pattern Recognition}, pages 1386--1395, 2017.

\bibitem{slavcheva2018sobolevfusion}
Miroslava Slavcheva, Maximilian Baust, and Slobodan Ilic.
\newblock Sobolevfusion: 3d reconstruction of scenes undergoing free non-rigid
  motion.
\newblock In {\em Proceedings of the IEEE conference on computer vision and
  pattern recognition}, pages 2646--2655, 2018.

\bibitem{robustfusion}
Zhuo Su, Lan Xu, Zerong Zheng, Tao Yu, Yebin Liu, and Lu Fang.
\newblock Robustfusion: Human volumetric capture with data-driven visual cues
  using a rgbd camera.
\newblock In Andrea Vedaldi, Horst Bischof, Thomas Brox, and Jan-Michael Frahm,
  editors, {\em Computer Vision -- ECCV 2020}, pages 246--264, Cham, 2020.
  Springer International Publishing.

\bibitem{su2022robustfusionPlus}
Zhuo Su, Lan Xu, Dawei Zhong, Zhong Li, Fan Deng, Shuxue Quan, and Lu Fang.
\newblock Robustfusion: Robust volumetric performance reconstruction under
  human-object interactions from monocular rgbd stream.
\newblock {\em IEEE Transactions on Pattern Analysis and Machine Intelligence},
  2022.

\bibitem{sumner2007embedded}
Robert~W Sumner, Johannes Schmid, and Mark Pauly.
\newblock Embedded deformation for shape manipulation.
\newblock {\em ACM Transactions on Graphics (TOG)}, 26(3):80, 2007.

\bibitem{sun2021HOI-FVV}
Guoxing Sun, Xin Chen, Yizhang Chen, Anqi Pang, Pei Lin, Yuheng Jiang, Lan Xu,
  Jingya Wang, and Jingyi Yu.
\newblock Neural free-viewpoint performance rendering under complex
  human-object interactions.
\newblock In {\em Proceedings of the 29th ACM International Conference on
  Multimedia}, 2021.

\bibitem{suo2021neuralhumanfvv}
Xin Suo, Yuheng Jiang, Pei Lin, Yingliang Zhang, Minye Wu, Kaiwen Guo, and Lan
  Xu.
\newblock Neuralhumanfvv: Real-time neural volumetric human performance
  rendering using rgb cameras.
\newblock In {\em Proceedings of the IEEE/CVF Conference on Computer Vision and
  Pattern Recognition}, pages 6226--6237, 2021.

\bibitem{suo2020neural3d}
Xin Suo, Minye Wu, Yanshun Zhang, Yingliang Zhang, Lan Xu, Qiang Hu, and Jingyi
  Yu.
\newblock Neural3d: Light-weight neural portrait scanning via context-aware
  correspondence learning.
\newblock In {\em Proceedings of the 28th ACM International Conference on
  Multimedia}, pages 3651--3660, 2020.

\bibitem{thies2019deferred}
Justus Thies, Michael Zollh{\"o}fer, and Matthias Nie{\ss}ner.
\newblock Deferred neural rendering: Image synthesis using neural textures.
\newblock {\em ACM Transactions on Graphics (TOG)}, 38(4):1--12, 2019.

\bibitem{tretschk2021nonrigid}
Edgar Tretschk, Ayush Tewari, Vladislav Golyanik, Michael Zollh\"{o}fer,
  Christoph Lassner, and Christian Theobalt.
\newblock Non-rigid neural radiance fields: Reconstruction and novel view
  synthesis of a dynamic scene from monocular video.
\newblock In {\em {IEEE} International Conference on Computer Vision ({ICCV})}.
  {IEEE}, 2021.

\bibitem{wang2022fourier}
Liao Wang, Jiakai Zhang, Xinhang Liu, Fuqiang Zhao, Yanshun Zhang, Yingliang
  Zhang, Minye Wu, Jingyi Yu, and Lan Xu.
\newblock Fourier plenoctrees for dynamic radiance field rendering in
  real-time.
\newblock In {\em Proceedings of the IEEE/CVF Conference on Computer Vision and
  Pattern Recognition}, pages 13524--13534, 2022.

\bibitem{wang2021neus}
Peng Wang, Lingjie Liu, Yuan Liu, Christian Theobalt, Taku Komura, and Wenping
  Wang.
\newblock Neus: Learning neural implicit surfaces by volume rendering for
  multi-view reconstruction.
\newblock {\em NeurIPS}, 2021.

\bibitem{wang2022reconstruction}
Xi Wang, Gen Li, Yen-Ling Kuo, Muhammed Kocabas, Emre Aksan, and Otmar
  Hilliges.
\newblock Reconstructing action-conditioned human-object interactions using
  commonsense knowledge priors.
\newblock In {\em International Conference on 3D Vision (3DV)}, 2022.

\bibitem{Wang_2020_WACV}
Zirui Wang, Shuda Li, Henry Howard-Jenkins, Victor Prisacariu, and Min Chen.
\newblock Flownet3d++: Geometric losses for deep scene flow estimation.
\newblock In {\em Proceedings of the IEEE/CVF Winter Conference on Applications
  of Computer Vision (WACV)}, March 2020.

\bibitem{weng_humannerf_2022_cvpr}
Chung-Yi Weng, Brian Curless, Pratul~P. Srinivasan, Jonathan~T. Barron, and Ira
  Kemelmacher-Shlizerman.
\newblock Human{N}e{RF}: Free-viewpoint rendering of moving people from
  monocular video.
\newblock In {\em Proceedings of the IEEE/CVF Conference on Computer Vision and
  Pattern Recognition (CVPR)}, pages 16210--16220, June 2022.

\bibitem{Wu_2020_CVPR}
Minye Wu, Yuehao Wang, Qiang Hu, and Jingyi Yu.
\newblock Multi-view neural human rendering.
\newblock In {\em Proceedings of the IEEE/CVF Conference on Computer Vision and
  Pattern Recognition (CVPR)}, June 2020.

\bibitem{xie2022chore}
Xianghui Xie, Bharat~Lal Bhatnagar, and Gerard Pons-Moll.
\newblock Chore: Contact, human and object reconstruction from a single rgb
  image.
\newblock In {\em European Conference on Computer Vision ({ECCV})}. {Springer},
  October 2022.

\bibitem{FlyFusion}
Lan Xu, Wei Cheng, Kaiwen Guo, Lei Han, Yebin Liu, and Lu Fang.
\newblock Flyfusion: Realtime dynamic scene reconstruction using a flying depth
  camera.
\newblock {\em IEEE transactions on visualization and computer graphics},
  27(1):68--82, 2019.

\bibitem{UnstructureLan}
Lan Xu, Zhuo Su, Lei Han, Tao Yu, Yebin Liu, and Lu Fang.
\newblock Unstructuredfusion: realtime 4d geometry and texture reconstruction
  using commercial rgbd cameras.
\newblock {\em IEEE transactions on pattern analysis and machine intelligence},
  42(10):2508--2522, 2019.

\bibitem{yoon2022learning}
Jae~Shin Yoon, Duygu Ceylan, Tuanfeng~Y Wang, Jingwan Lu, Jimei Yang, Zhixin
  Shu, and Hyun~Soo Park.
\newblock Learning motion-dependent appearance for high-fidelity rendering of
  dynamic humans from a single camera.
\newblock In {\em Proceedings of the IEEE/CVF Conference on Computer Vision and
  Pattern Recognition}, pages 3407--3417, 2022.

\bibitem{yu2021plenoctrees}
Alex Yu, Ruilong Li, Matthew Tancik, Hao Li, Ren Ng, and Angjoo Kanazawa.
\newblock Plenoctrees for real-time rendering of neural radiance fields.
\newblock In {\em Proceedings of the IEEE/CVF International Conference on
  Computer Vision}, pages 5752--5761, 2021.

\bibitem{BodyFusion}
Tao Yu, Kaiwen Guo, Feng Xu, Yuan Dong, Zhaoqi Su, Jianhui Zhao, Jianguo Li,
  Qionghai Dai, and Yebin Liu.
\newblock Bodyfusion: Real-time capture of human motion and surface geometry
  using a single depth camera.
\newblock In {\em The IEEE International Conference on Computer Vision (ICCV)}.
  ACM, October 2017.

\bibitem{yu2021function4d}
Tao Yu, Zerong Zheng, Kaiwen Guo, Pengpeng Liu, Qionghai Dai, and Yebin Liu.
\newblock Function4d: Real-time human volumetric capture from very sparse
  consumer rgbd sensors.
\newblock In {\em Proceedings of the IEEE/CVF Conference on Computer Vision and
  Pattern Recognition}, pages 5746--5756, 2021.

\bibitem{DoubleFusion}
Tao Yu, Zerong Zheng, Kaiwen Guo, Jianhui Zhao, Qionghai Dai, Hao Li, Gerard
  Pons-Moll, and Yebin Liu.
\newblock Doublefusion: Real-time capture of human performances with inner body
  shapes from a single depth sensor.
\newblock {\em Transactions on Pattern Analysis and Machine Intelligence
  (TPAMI)}, 2019.

\bibitem{zhang2020phosa}
Jason~Y. Zhang, Sam Pepose, Hanbyul Joo, Deva Ramanan, Jitendra Malik, and
  Angjoo Kanazawa.
\newblock Perceiving 3d human-object spatial arrangements from a single image
  in the wild.
\newblock In {\em European Conference on Computer Vision (ECCV)}, 2020.

\bibitem{zhang2020object}
Tianshu Zhang, Buzhen Huang, and Yangang Wang.
\newblock Object-occluded human shape and pose estimation from a single color
  image.
\newblock In {\em Proceedings of the IEEE/CVF Conference on Computer Vision and
  Pattern Recognition}, pages 7376--7385, 2020.

\bibitem{zhao2022human}
Fuqiang Zhao, Yuheng Jiang, Kaixin Yao, Jiakai Zhang, Liao Wang, Haizhao Dai,
  Yuhui Zhong, Yingliang Zhang, Minye Wu, Lan Xu, and Jingyi Yu.
\newblock Human performance modeling and rendering via neural animated mesh.
\newblock {\em ACM Trans. Graph.}, 41(6), nov 2022.

\bibitem{zhao2022humannerf}
Fuqiang Zhao, Wei Yang, Jiakai Zhang, Pei Lin, Yingliang Zhang, Jingyi Yu, and
  Lan Xu.
\newblock Humannerf: Efficiently generated human radiance field from sparse
  inputs.
\newblock In {\em Proceedings of the IEEE/CVF Conference on Computer Vision and
  Pattern Recognition}, pages 7743--7753, 2022.

\bibitem{zheng2022structured}
Zerong Zheng, Han Huang, Tao Yu, Hongwen Zhang, Yandong Guo, and Yebin Liu.
\newblock Structured local radiance fields for human avatar modeling.
\newblock In {\em Proceedings of the IEEE/CVF Conference on Computer Vision and
  Pattern Recognition}, pages 15893--15903, 2022.

\bibitem{HybridFusion}
Zerong Zheng, Tao Yu, Hao Li, Kaiwen Guo, Qionghai Dai, Lu Fang, and Yebin Liu.
\newblock Hybridfusion: Real-time performance capture using a single depth
  sensor and sparse imus.
\newblock In {\em European Conference on Computer Vision (ECCV)}, Sept 2018.

\end{thebibliography}
}

\end{document}